\documentclass[11pt]{article}

\usepackage[final]{acl}

\usepackage{times}
\usepackage{latexsym}
\usepackage{xcolor}
\definecolor{dullorange}{RGB}{204,119,34}

\usepackage{geometry}
\usepackage{tabularx}
\usepackage{booktabs}  
\usepackage{xcolor}    
\usepackage{array}     
\usepackage{lipsum}    
\usepackage{fancyvrb}
\usepackage{fvextra} 
\usepackage{calc}    
\usepackage{graphicx}
\usepackage{longtable}
\usepackage{multirow}
\usepackage{amssymb}
\usepackage{amsfonts}
\usepackage{minted}
\usepackage{amsmath}

\DeclareMathOperator*{\argmin}{argmin}

\newcommand{\e}{\ensuremath{ \mbox{\scriptsize{E}} }}

\usepackage[T1]{fontenc}

\usepackage[utf8]{inputenc}

\usepackage{microtype}

\usepackage{inconsolata}

\usepackage{graphicx}

\usepackage{bm}
\usepackage{listings}
\usepackage{xcolor}

\definecolor{systemprompt}{RGB}{0, 102, 204}      
\definecolor{userprompt}{RGB}{0, 153, 76}         
\definecolor{formatting}{RGB}{204, 51, 0}         

\lstdefinestyle{promptstyle}{
  basicstyle=\small\ttfamily,
  breaklines=true,
  frame=single,
  xleftmargin=0.5cm,
  xrightmargin=0.5cm,
  moredelim=[is][\color{systemprompt}\bfseries]{<SYS>}{</SYS>},
  moredelim=[is][\color{userprompt}\bfseries]{<USER>}{</USER>},
  moredelim=[is][\color{formatting}\bfseries]{<FMT>}{</FMT>}
}

\title{LLM-HYPER: Generative CTR Modeling for Cold-Start Ad Personalization via LLM-Based Hypernetworks}

\author{
  Luyi Ma\thanks{Equal contribution.} \quad 
  Wanjia Sherry Zhang\footnotemark[1] \quad
  Zezhong Fan\footnotemark[1] \quad
  Shubham Thakur\footnotemark[1] \quad 
  Kai Zhao \\
  \textbf{Kehui Yao} \quad 
  \textbf{Ayush Agarwal} \quad 
  \textbf{Rahul Iyer} \quad 
  \textbf{Jason Cho} \\
  \textbf{Jianpeng Xu} \quad 
  \textbf{Evren Korpeoglu} \quad 
  \textbf{Sushant Kumar} \quad 
  \textbf{Kannan Achan} \\
  \text{Walmart Global Tech} \\
  Sunnyvale, California, USA \\
  \texttt{\{luyi.ma, sherry.zhang, zezhong.fan, shubham.thakur0, kai.zhao\}@walmart.com} \\
  \texttt{\{kehui.yao, Ayush.Agarwal, rahul.iyer, jason.cho\}@walmart.com} \\
  \texttt{\{jianpeng.xu, EKorpeoglu, sushant.kumar, kannan.achan\}@walmart.com}
}

\begin{document}
\maketitle
\begin{abstract}
On online advertising platforms, newly introduced promotional ads face the cold-start problem, as they lack sufficient user feedback for model training. In this work, we propose LLM-HYPER, a novel framework that treats large language models (LLMs) as hypernetworks to directly generate the parameters of the click-through rate (CTR) estimator in a training-free manner. LLM-HYPER
uses few-shot Chain-of-Thought prompting over multimodal ad content (text and images) to infer feature-wise model weights for a linear CTR predictor. By retrieving semantically similar past campaigns via CLIP embeddings and formatting them into prompt-based demonstrations, the LLM learns to reason about customer intent, feature influence, and content relevance. To ensure numerical stability and serviceability, we introduce normalization and calibration techniques that align the generated weights with production-ready CTR distributions. Extensive offline experiments show that LLM-HYPER significantly outperforms cold-start baselines in NDCG$@10$ by 55.9\%. Our real-world online A/B test on one of the top e-commerce platforms in the U.S. demonstrates the strong performance of LLM-HYPER, which drastically reduces the cold-start period and achieves competitive performance. LLM-HYPER has been successfully deployed in production.
\end{abstract}

\section{Introduction}
In the rapidly evolving landscape of online advertising, platforms face a persistent `cold-start' challenge: newly introduced promotional ads often lack sufficient historical user feedback for effective training of CTR ranking models~\cite{park2009pairwise, lee2019melu, lu2020meta, wei2021contrastive}, for example, during rapid transitions from Christmas to New Year promotions. Improper cold-start solutions could significantly harm ad relevance and platform revenue within the limited conversion window.  

High-stakes advertising environments, such as e-commerce homepages, demand not only predictive accuracy but also operational transparency and controllability to satisfy business policies (e.g., brand and price) and audit requirements. In industrial CTR ranking pipelines, linear layers are widely used to aggregate high-level user preference representations from low-level interaction histories and calibrate signals from various channels for final CTR estimation \cite{mcmahan2013ad, cheng2016wide, Liang2020JTCN, WangYXSDZ22, yan2022scale}, owing to their structural simplicity. 
Because their coefficients directly measure feature contributions, linear layers further provide the built-in transparency needed to meet regulatory and policy standards.
This makes an interpretable linear scoring function a desirable control component when integrated with complex deep or LLM-based models, helping to satisfy explanation requirements and mitigate risks associated with opaque model behavior \cite{Roustaei2024LinearRegression}.





While Large Language Models (LLMs) have demonstrated remarkable semantic reasoning capabilities that could directly generate the target ranking in natural language with reasoning and explanation \cite{geng2022recommendation, cui2022m6, liao2023llara, tan2024idgenrec, ma2024triple} by contextualizing user and item representations~\cite{qiu2021u, yao2022reprbert, zhang2022gbert, hou2022towards, li2023text}, their direct deployment in real-time ranking environments is frequently prohibitive. Industrial ranking systems operate under extreme micro-latency constraints, often requiring low p99 response times. Such rigorous benchmarks effectively preclude the direct of heavy and costly architectures or real-time LLM inference for every ranking request. 

Our paper focuses on the critical research question: How to leverage LLMs' reasoning capacity to derive a linear ranking solution without training labels (cold start) while decoupling costly LLM inference from the low-latency, fast-evolving deployment environment at an industrial scale (efficient deployment).



To address this research question, we present LLM-HYPER, a novel framework deployed in production that leverages Large Language Models (LLMs) as hypernetworks~\cite{ha2016hypernetworks, chauhan2024brief} to generate linear model weights for cold-start ads ranking in a training-free manner. 
LLM-HYPER prompts an LLM (e.g., Gemini-2.5) with the text and image content of the cold-start ads (e.g., title, image, and metadata) \cite{he2017neural, acharya2023llm}, and the definition of user features to directly infer feature-wise weights for a ready-to-deploy linear ranking model, which is different from existing hypernetwork solutions with strict dependency of training data \cite{zhang2018graph, knyazev2021parameter, zhmoginov2022hypertransformer, beck2023hypernetworks, knyazev2023can, zhang2024metadiff}. 

Specifically, we retrieve warm ads similar to the cold-start ad and use their text, image, and trained model weights as few-shot examples for Chain-of-Thought (CoT)~\cite{wei2022chain} reasoning in the weight-generation prompt.
Since user features will be used to multiply the generated model weights to compute the ranking score, we include the user feature definition in the prompt to reduce the context gap between the generation stage (NLP reasoning) and the model-serving stage (Numeric computation). We define generation criteria for LLM reasoning and reflection~\cite{shinn2023reflexion}. 

\textbf{Industrial Deployment}: LLM-HYPER decouples the time-consuming weight generation by LLM from scalable serving. Before the launch date of cold-start ads, LLM-HYPER has sufficient time to generate the model weights offline. A label-independent normalization and calibration mechanism is subsequently introduced to stabilize model weights for robust deployment, ensuring that the generated weights yield fast, numerically stable, and well-calibrated CTR predictions in production. 
Results from offline evaluation on the proprietary interaction data show that LLM-HYPER significantly outperforms cold-start baselines. LLM-HYPER achieves convincing CTR performance comparable to the in-production warm-start model in a 30-day online A/B deployment on one of the top e-commerce platforms in the U.S. LLM-HYPER is successfully deployed in production to power the Homepage Ads cold-start ranking. 

\begin{figure*}[t]
    \centering
    \includegraphics[width=0.90
\linewidth]{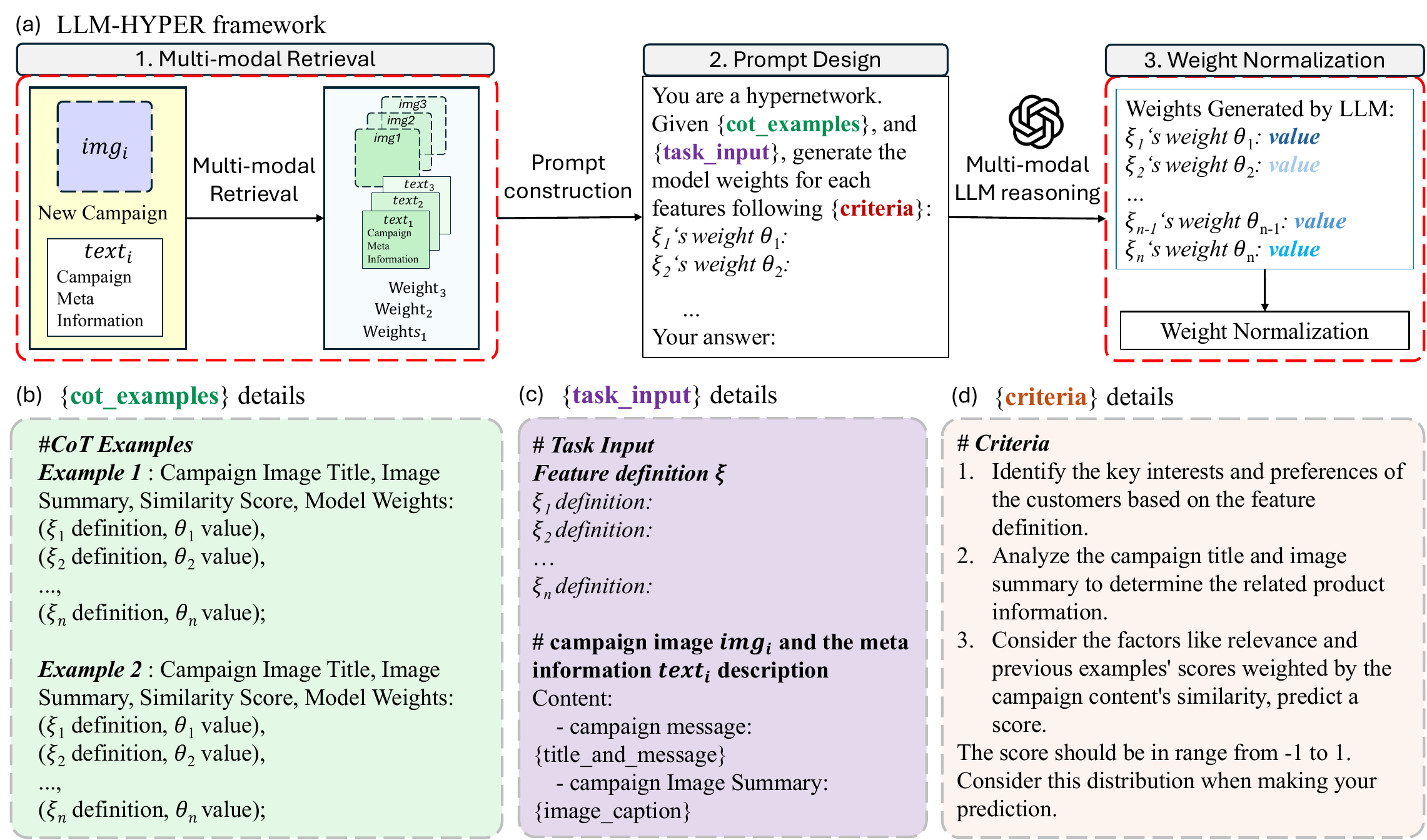}
    \caption{LLM-HYPER: (a) framework and (b)-(d) prompt details. Weights are normalized for CTR prediction. 
    }
    \label{fig:main_feature}
\end{figure*}

\section{Methodology}
In this section, we describe the details of LLM-HYPER for weight generation. We first present the prompt details for LLM reasoning and weight generation, and explain multimodal retrieval for Chain-of-Thought (CoT) few-shot learning. Finally, we address the normalization of model weights for production-ready CTR prediction. Figure \ref{fig:main_feature} illustrates the framework. 

\label{sec:method}
\subsection{Preliminaries: Linear CTR Estimator}
Let $\mathcal{U}$ be the set of customers with $|\mathcal{U}| = N$ and $\mathcal{R}$ be the set of ads with $|\mathcal{R}| = M$.  
For a customer $u \in \mathcal{U}$ and a recommended ad $r \in \mathcal{R}$,  
let the $n$-dimensional personalized feature vector be $\bm{\xi}^u = \bigl[\xi^u_{0}, \xi^u_{1}, \ldots, \xi^u_{n}\bigr]$, constructed from historical interactions and contextual signals.  Denote the weight vector for an ad $r$ as $\bm{\theta}^r = \bigl[\theta^r_{0}, \theta^r_{1}, \ldots, \theta^r_{n}\bigr]$,
the predicted click‐through rate (CTR) for the pair $(u,r)$ is $p(u,r) = \sigma\bigl\{(\bm{\theta}^r)' \,\bm{\xi}^u\bigr\}$, where $\sigma(x) = \bigl(1 + e^{-x}\bigr)^{-1}$ is the sigmoid function. In the traditional supervised learning setting, each $\bm{\theta}^r$ is learned by minimizing the log‐loss over all users:

\begin{align*}
\mathcal{L} = - \frac{1}{N} \sum_{u} &\Big[ y_r^u \log p(u,r) \\
             &+ (1 - y_r^u) \log\{1 - p(u,r)\} \Big]
\end{align*}
where $y_r^u$ is the binary label indicating whether user $u$ clicked on ad $r$.

In LLM-HYPER, the weight vector $\bm{\theta}^r$ for each cold ad $r$ is directly generated by LLM.
Formally, given $\text{prompt}_{\bm{\xi},r}
= \bigl[\text{text}_{\bm{\xi}},\; \text{img}_r,\; \text{text}_{r},\; \text{few-shot}_{r}\bigr],$
and then generate $\bm{\theta}^r = \mathrm{LLM}\bigl(\text{prompt}_{\bm{\xi},r}\bigr).$ We explain the prompt formulation in section \ref{section:prompt-design}.
Note that this linear model generalizes to an internal feature aggregation layer in complex deep models~\cite{Liang2020JTCN, WangYXSDZ22} and to the linear calibration layer over different signal channels~\cite{yan2022scale}, which is beyond the scope of this paper. We focus on demonstrating the essential step of LLM-generated linear weights, specifically under the CTR prediction task.


\subsection{Prompt Design}\label{section:prompt-design}
Our prompt template contains three essential parts: \texttt{cot\_example} for contextualization, \texttt{task\_input} for cold ad materialization, and \texttt{criteria} for generation guidance and reflection.

\noindent \textbf{\texttt{cot\_example}}: \label{section:mm-rag}
Ads involve both rich text information (e.g., titles and messages) and visual information (images) to attract customers' attention. To find informative past ads for few-shot CoT, we use CLIP \cite{radford2021learning} to encode the ads text and image data into embeddings, $e_i = \text{CLIP}(\text{img}_i, \text{text}_i)$, and build a multimodal retrieval\footnote{KNN by Faiss: https://github.com/facebookresearch/faiss} based on that, 
\begin{equation}
    \text{KNN}_r = \argmin_{i \in H, |S|=k}  \sum_{i\in S} \|e_i - e_r\|,
    \label{eq:mm-retrieval}
\end{equation} where $H$ is a set of past campaigns, $e_r$ is the CLIP embedding of the current candidate $r$. 
With the indexes of top-K ads $\text{KNN}_r$, we retrieve their text, images, meta information and also past trained weight vector as few-shot examples, $\text{few-shot}_r = \{(\text{text}_i, \text{img}_i, \bm{\theta}^i) | i \in \text{KNN}_R\}$, where $\bm{\theta}^i$ is trained using traditional linear methods based on past warm ad data (Figure \ref{fig:main_feature}-a(1)). These few-shot examples, together with the current campaign $(\text{text}_r, \text{img}_r)$, are used for prompt formulation (Figure \ref{fig:main_feature}-b).
We provide the similarity score between the selected examples and the current new candidate for the LLM to transfer  knowledge.

\noindent \textbf{\texttt{task\_input} and \texttt{criteria}}: The instruction section of the prompt emphasizes the definition of input user features and cold ad's text and visual caption (generated by LLMs) for LLM to reason the target network weights (Figure \ref{fig:main_feature}-c). We provide weight generation criteria (Figure \ref{fig:main_feature}-d) as guidance and reflection. 
We specify the output format by explicitly requiring one weight per line per feature to guide the weight generation (Appendix \ref{app:llm_hyper_result_visualiation} \& \ref{app:llm_hyper_prompt_design}).

\subsection{Weight Normalization}
To enhance the numeric stability, we first normalize the predicted weight vector $\bm{\theta}^r$, producing a new parameter vector $ \tilde{\bm{\theta}^r} = \bm{\theta}^r/\lvert\lvert \bm{\theta}^r \rvert\rvert$. This step mitigates the risk of overfitting by constraining large parameter values. 

We further calibrate the normalized weight vector by shifting, which aligns the distribution of its predicted probabilities with those of pre-existing, fully trained models and ensures the seamless integration into the online serving environment. 
Specifically, the intercept term is shifted by $\delta$ so that,
\begin{equation}
     \mathbb{E}[\sigma\{(\tilde{\bm{\theta}^r})^{'} \bm{\xi}^u + \delta\}] = \alpha, 
     \label{eq:intercept}
\end{equation} where $\alpha$ is the average predicted probability of similar available contents in Equation \ref{eq:mm-retrieval}. The value of the intercept shift $\delta$ can be derived from Equation \ref{eq:intercept} by computation numerically. 
This adjustment can be approached by sampling customer data to get the distribution of $(\tilde{\bm{\theta}^r})^{'} \bm{\xi}^u$, so that final output probabilities align with the intended balance of positive and negative predictions. Our LLM-HYPER only needs to run weight prediction once before pushing to online exposure without time-consuming label collection, streamlining the model definition and launch.

\subsection{LLM-HYPER Deployment}
\begin{figure}[t]
    \centering
    \includegraphics[width=\linewidth]{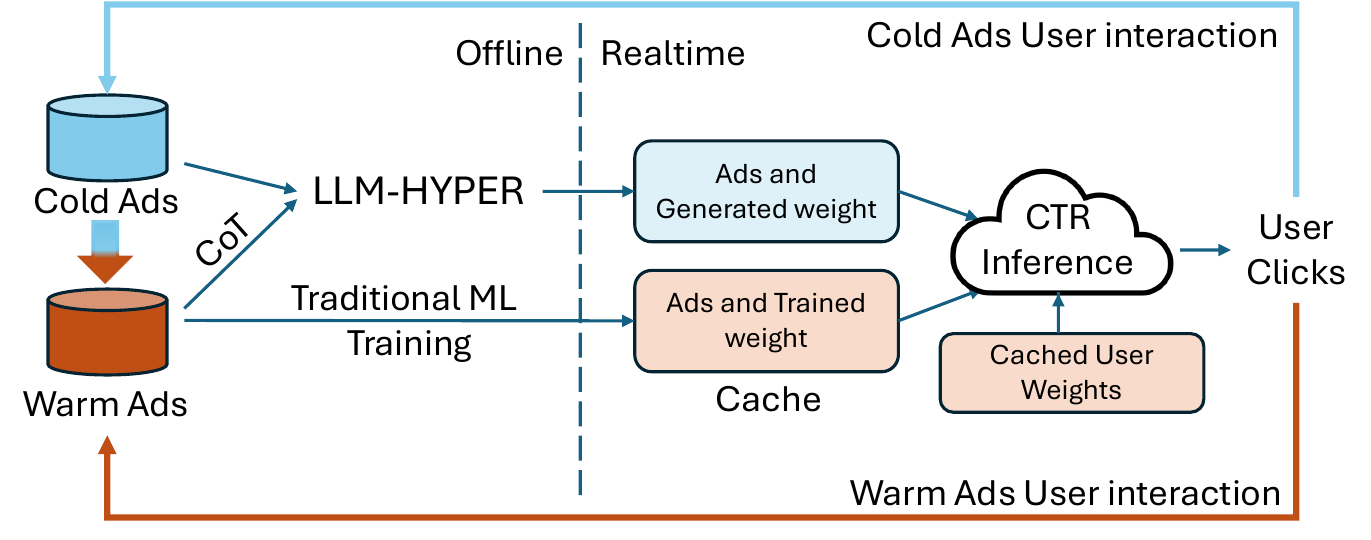}
    \caption{LLM-HYPER deployment}
    \label{fig:deployment}
\end{figure}

When new cold ads $R_{cold} = \{r_1, r_2, ...\}$ come in, LLM-HYPER generates model weights $\Theta = \{\bm{\theta}_1, \bm{\theta}_2,... \}$ for each cold ad for CTR estimation in offline environment before the ad launch date. The generated weights could be stored in a cache accessible in real time or on a server for real-time ranking inference, thereby overcoming the cold-start issue. 
After launching the cold-start ads, cold ads are warmed up, and interaction data are collected to fuel the traditional machine learning-based CTR model training. The platform then use trained CTR model for warm-start ranking inference (Figure \ref{fig:deployment}). Appendix \ref{app:deployment_insights} provides additional deployment insights. 

\section{Experiments}
We conduct experiments with the aim of answering the following research questions: \textbf{RQ1}: The effectiveness of LLM-HYPER for weight generation and CTR ranking in the offline environment; \textbf{RQ2}: Ablation study on the impact of CoT and visual features for weight generation; \textbf{RQ3}: Explainability of LLM-HYPER for weight generation; \textbf{RQ4}: Robustness of LLM-HYPER for weight generation.

\subsection{Dataset Description} 
To evaluate models effectively offline, we randomly sample 675 warm ads and user-interaction feedback from the past 3 months, involving 1,000,000 users, on one of the largest e-commerce platforms in the U.S. 
To simulate the cold-start scenario offline, we first partition the dataset chronologically: 455 earlier ads and their trained model weights constitute the \textit{retired} ad set, and the remaining 120 more recent ads constitute the \textit{active} ad set. 
The \textit{active} ad set is further split into training and test sets: interactions of 800,000 users are used for training, and interactions from 200,000 users are held out for testing. 
Due to the proprietary nature of the dataset, further details are not disclosed.

\subsection{Baselines}
\noindent \textbf{Warm-start Baseline}: \textbf{LR$_{warm}$} represents the traditional "warm-start" scenario where a linear CTR model is trained on the training set of 120 active ads and evaluated on the test set. This sets an ideal case for new ads though it is typically unattainable in a true cold-start setting.

\noindent \textbf{Cold-start Baselines}: 
We consider three types of cold-start baselines. First is \textbf{Emb$_{T5}$}, a popular embedding-based cold-start recommender baseline that computes the cosine similarity between the user and ad Sentence-T5-base~\cite{ni2022sentence} semantic embeddings, built from their respective NLP contexts. The Ad NLP context is built by the LLM from the title and image caption, whereas the user's NLP context is formed by the LLM summarizing the user's past ad interactions. 
Second, we consider \textbf{LLM-R}~\cite{hou2024llmrank} and \textbf{LLM-TR}~\cite{zhang2025llmtreerec} as LLM recommender baselines for cold-start scenarios. These LLM recommenders rank ads based on user and ad NLP contexts.
Finally, we consider a heuristic baseline for the low-latency requirements.
\textbf{LR$_{cold}$}: The model weights are set to the median values of weights across all 455 \textit{retired} ads to mitigate the influence of outlier ads.
All baseline models are implemented using the reported configuration and parameters. 

\noindent \textbf{Cold-start LLM-HYPER (ours)}: We create variants of LLM-HYPER based on the types of LLM used. \textbf{LH$_{2.5P}$} and \textbf{LH$_{2.5F}$} use Gemini-2.5-Pro and 2.5-Flash~\cite{comanici2025gemini}, respectively.  \textbf{LH$_{4o}$} uses GPT-4o~\cite{hurst2024gpt} and \textbf{LH$_{5.1}$} uses GPT-5.1\footnote{https://openai.com/index/gpt-5-1/}. We use 5-shot CoT and $temperature=0.5$ for all LLM-HYPER variants (see Appendix \ref{app:llm_hyper_implement} for additional details).

We report AUC and NDCG$@\{5, 10\}$ for offline ranking performance on the test set, and relative CTR score for online A/B testing. We report serving-time efficiency using the average latency metric \textbf{L}(ms) for CTR prediction across all 120 active ads.



\subsection{Experimental Results}
We present offline evaluation results in Table \ref{tab:model-performance}. 
\noindent \textbf{Performance}: While \textbf{LR$_{warm}$} sets the upper bound of the performance due to its ideal but unattainable setting in the cold start, all LLM-HYPER baselines outperform other cold-start baselines (p-value $\leq 0.05$), with \textbf{LH$_{2.5P}$} achieving the best cold-start AUC (+20.2\%) and NDCG$@10$ (+55.9\%) compared with the best cold-start baseline \textbf{LR$_{cold}$}. 
\textbf{Emb$_{T5}$} does not perform well in strict cold-start scenarios due to the big context gap between offline ad relevance and online user feedback~\cite{song2024importance}, while LLM-HYPER overcomes this gap by effective LLM reasoning. 
\textbf{LLM-R} and \textbf{LLM-TR} have the worst performance over the metrics due to the hallucinated and out-of-distribution prediction~\cite{jiang2025beyond}. For example, these two LLM recommenders might generate an inaccurate ad name or index without further correction, and rank too novel ads with zero interaction to top positions. These issues hurt their performance and applicability. LLM-HYPER has fewer issues with ad name hallucination due to its direct weight generation and normalization. 

\noindent \textbf{Latency}: LLM-HYPER variants achieve the same competitive latency as other linear baselines within a range of $(0.14, 0.17)$ milliseconds on average, generating a ranking for a user. This low latency is because we decouple the costly LLM inference steps from the serving stage, only inferencing the linear computation. 
LLM recommender baselines take much more time to respond due to time-consuming LLM calls. \textbf{LLM-TR} achieves better performance than \textbf{LLM-R} at the cost of additional hierarchical LLM calls, further increasing the latency. LLM-HYPER offers a great balance between costly LLM reasoning and low-latency serving (\textbf{RQ1}).

\subsection{Ablation Study: CoT and Visual Feature}
To study the impact of CoT and visual feature input on performance, we conducted a series of ablation studies that varied the number of few-shot examples and the exclusion of image information from the prompt on the best LLM-HYPER variant \textbf{LH$_{2.5P}$}. Results are summarized in Table \ref{tab:ablation}. 
All LLM-HYPER variants with visual features outperform \textbf{LR$_{cold}$}, the best cold-start baseline in Table \ref{tab:model-performance}. Even the simplest zero-shot setting achieves a 33.3\% improvement in NDCG$@10$, demonstrating strong cross-modal generalization. The 5-shot configuration achieves the best balance across all ranking metrics, though the 3-shot configuration shows a slight advantage in NDCG$@10$, suggesting that a moderate number of high-quality examples is sufficient to learn the weight-distribution logic.
Excluding visual features from the prompt drastically reduces performance across all variants, indicating the crucial role of visual content understanding in the LLM reasoning and user-ad interaction (\textbf{RQ2}). 

\begin{table}[htbp]
\centering
\small
\begin{tabular}{lcccc}
\toprule
\textbf{Model} & \textbf{AUC} & \textbf{NDCG5} & \textbf{NDCG10} & \textbf{L(ms)}\\
\midrule
\textbf{Emb$_{T5}$} & 0.4105  & 0.0174 & 0.0262 & 0.357\\
\midrule
\textbf{LLM-R} & 0.0571 & 0.00931 & 0.0221 & $2.13\e{3}$\\
\textbf{LLM-TR} & 0.0576 & 0.00886 & 0.0508 & $3.78\e{3}$ \\
\midrule
\textbf{LH$_{4o}$}         & 0.6539 & \underline{0.0490} & 0.0764 & 0.163\\
\textbf{LH$_{5.1}$}        & 0.6497 & 0.0427 & 0.0668 & 0.158\\
\textbf{LH$_{2.5F}$} & 0.6654 & 0.0435 & 0.0779 & 0.160\\
\textbf{LH$_{2.5P}$}   & \underline{0.6722} & 0.0456 & \underline{0.0792} &  0.157\\ 
\midrule
\textbf{LR$_{cold}$} & 0.5593 & 0.0328 & 0.0508  & 0.147 \\
\textbf{LR$_{warm}$}
& \textbf{0.7005} & \textbf{0.0639} & \textbf{0.0871} & 0.151 \\
\bottomrule
\end{tabular}
\caption{Offline Cold-start Ranking Comparison.}
\label{tab:model-performance}
\end{table}

\begin{table}[h!]
\centering
\small
\begin{tabular}{l c c}
\toprule
\textbf{Models} & \textbf{NDCG5} & \textbf{NDCG10} \\ \hline
\textbf{LH$_{2.5P}$}, zero-shot & 0.0376 & 0.0677 \\ 
\textbf{LH$_{2.5P}$}, zero-shot, $- img$ & 0.0165 & 0.0352\\ \hline
\textbf{LH$_{2.5P}$}, 3-shot & 0.0421 & \textbf{0.0798} \\ 
\textbf{LH$_{2.5P}$}, 3-shot, $- img$ & 0.0267 & 0.0511 \\ \hline
\textbf{LH$_{2.5P}$}, 5-shot & \textbf{0.0456} & 0.0792 \\ 
\textbf{LH$_{2.5P}$}, 5-shot, $- img$ & 0.0292 & 0.0506 \\ 
\bottomrule
\end{tabular}
\caption{CoT and Visual Feature Impact}
\label{tab:ablation}
\end{table}


\subsection{Explainability}




Explainability is essential for high-stakes advertising environments. Because LLM-HYPER bypasses the traditional model training process to directly generate model weights, it is critical to assess the explainability and trustworthiness of the feature importance generated by LLMs. 

\noindent \textbf{Human-labeled Dataset}: Since weights generated by LLM approximate the importance of user preference, we asked marketing experts to build an explainable ground-truth dataset on the feature importance. Market experts identify the most important user features (one or multiple) for each ad in the test set. This identified feature serves as the ground truth for evaluating whether LLMs' reasoning and outputs capture these ground-truth labels.

\noindent \textbf{Evaluation Metrics:} HitRate$@5$ (\textbf{HR$@5$}) to measure the probability of the ground-truth features appearing in the top-5 most important features reflected by LLM-generated weights. \textbf{Coverage}$@5$ to measure the extent to which the top-5 predicted preferences recover the attribute space of the ground-truth features. Additionally, we compute \textbf{Consistency Rate} to assess the logical agreement between the LLM's natural-language reasoning and its numerical weight outputs with respect to the ground truth. These metrics collectively quantify the model's ability to map ad features to human-annotated labels for explainability (see Appendix \ref{app:appendix_metrics} for detailed definitions of the metrics).

\begin{figure}[ht]
\centering
\includegraphics[width=0.96\columnwidth]{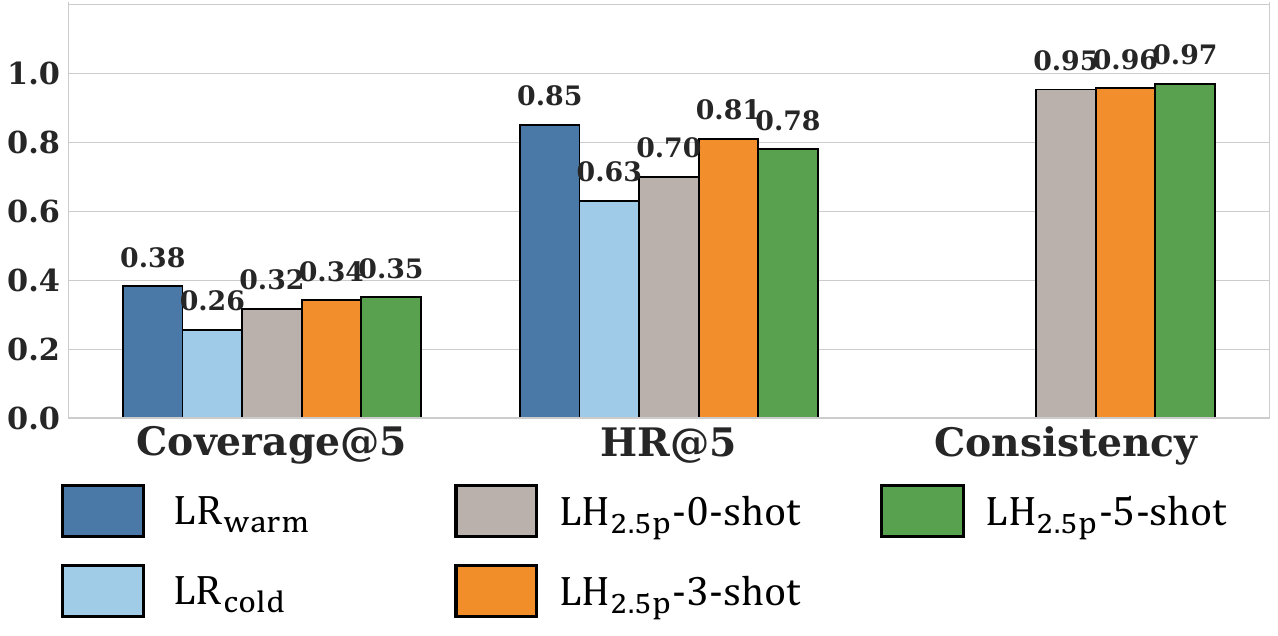}
\caption{Explainability Results of Cold Start Model}
\label{fig:explainability}
\vspace{-0.5cm}
\end{figure}

\noindent \textbf{Explainability and Alignment Results:} Figure~\ref{fig:explainability} shows that increasing the number of shots directly enhances alignment with target features. Coverage$@5$ improves from $0.317$ (zero-shot) to $0.351$ (5-shot), while HR$@5$ hits its maximum at $0.81$ in the 3-shot setting. This indicates that CoT examples serve as semantic anchors, helping the LLM accurately leverage predicted weights to the intended features. Furthermore, the Consistency score remains remarkably high ($>0.95$) across all settings, confirming that predicted weights are grounded in explicit, human-interpretable reasoning. This high level of instruction-following ensures the weight scale faithfully reflects the preferences identified in the reasoning phase, fulfilling industrial transparency requirements (\textbf{RQ3}).

\subsection{Robustness of LLM-HYPER}
In industrial applications, ad ranking must be robust against semantic perturbations and adversarial modifications, especially when ground-truth labels are unavailable for real-time fine-tuning. We evaluate this robustness by applying three types of counterfactual modifications to ads: \textit{Enhanced}, \textit{Diminished}, and \textit{Neutralized}. In \textit{Enhanced}, LLMs rewrite the ad title to enhance human-labeled features (e.g., by adding more information about dogs). In \textit{Diminished}, LLMs rewrite the ad title, thereby conflicting with human-labeled features (e.g., by switching from dog to cat information). In \textit{Neutralized}, LLM neutralizes the human-labeled features in the ad title (e.g., by switching from dog to just pets). We evaluate the accuracy of the binary classification task to determine whether LLM-HYPER changes the generated weights for the human-labeled features in the same semantic direction as the modification (see Appendix \ref{app:counterfactual} for more details).

\begin{figure}[ht]
\centering\includegraphics[width=0.96\columnwidth]{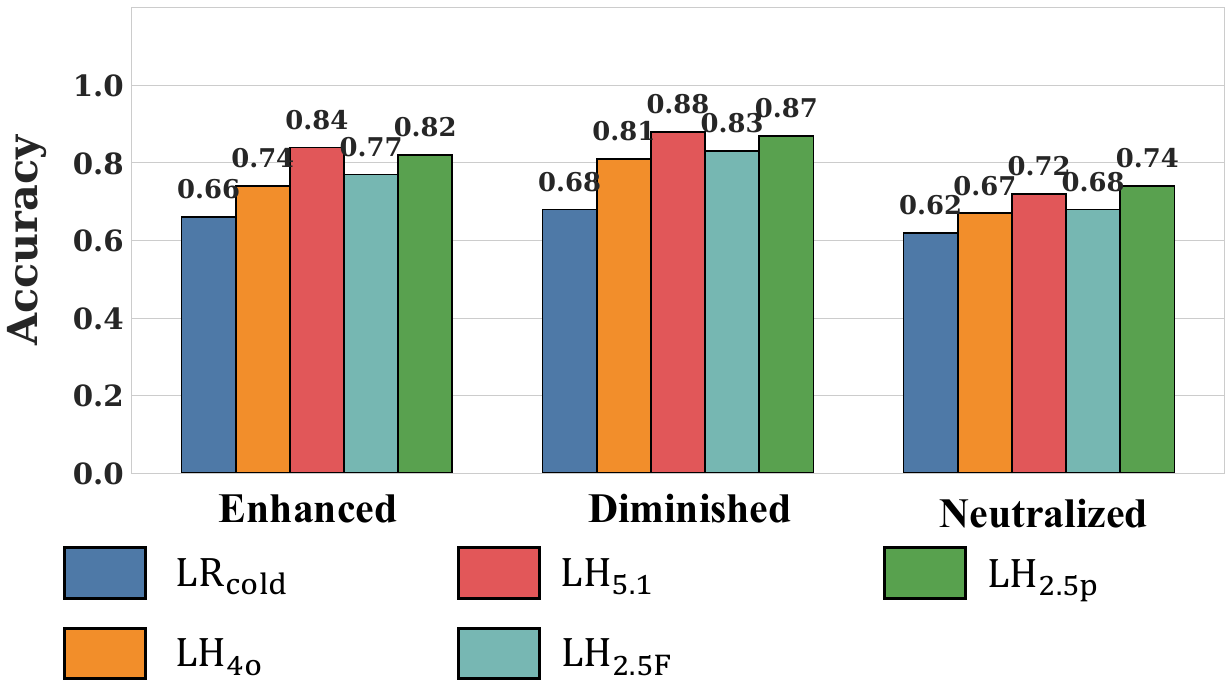}
\caption{Counterfactual Robustness of Cold Start Model with accuracy of weight change}
\label{fig:robustness}
\vspace{-0.5cm}
\end{figure}

\noindent \textbf{Results}: As illustrated in Figure~\ref{fig:robustness}, our framework exhibits significant robustness under cold start scenarios. GPT-5.1 (\textbf{LH}${_{5.1}}$) demonstrates the context reasoning capabilities, achieving an accuracy of $0.88$ in the \textit{Diminished} setting, effectively penalizing weights that deviate from the target features. While all backbones (including GPT-4o \textbf{LH}${_{4o}}$ and Gemini 2.5 series \textbf{LH}${_{2.5F/P}}$) show a performance decline in the \textit{Neutral} category, the overall trend proves that LLM-HYPER can reliably adjust weights based on semantic shifts. This capability is crucial for industrial deployment, as it allows the system to reflect preference changes through reasoning-weight consistency rather than relying on historical interaction data. LLM-HYPER demonstrates promising cold-start robustness that captures semantic changes while maintaining full interpretability for human auditors (\textbf{RQ4}).



\section{Online A/B Test and Deployment}
We conducted a 30-day online A/B test on the Homepage Ad ranking to provide an end-to-end, industrial-scale evaluation. 
To establish a baseline, we consider \textbf{LR}$_{warm}$ as the control, with pre-collected user feedback used for model training before the A/B test. 
Model weights generated by LLM-HYPER are in the variant group. 
The active ads are not exposed to the variant group for strict cold-start scenarios. Table \ref{tab:ab-test} summarizes the results. 

\begin{table}[h!]
\centering
\small
\begin{tabular}{lc}
\toprule
\textbf{Models} & \textbf{Relative CTR score} \\ 
\midrule
\textbf{LLM-HYPER}  & 92\%\\ 
\textbf{LR}$_{warm}$  & 100\% \\ 
\bottomrule
\end{tabular}
\caption{Online 30-Day A/B Test Results.}
\label{tab:ab-test}
\end{table}

LLM-HYPER achieves competitive CTR performance with the ideal warm-start \textbf{LR}$_{warm}$, with no statistically significant difference observed (p-value = 0.62), demonstrating the effectiveness of LLM-HYPER in generating weights for strict cold-start in a production environment.  
LLM-HYPER has been successfully deployed in production.

\bibliography{custom}

@String(AAAI = {AAAI})

@inproceedings{wei2021contrastive,
  title={Contrastive learning for cold-start recommendation},
  author={Wei, Yinwei and Wang, Xiang and Li, Qi and Nie, Liqiang and Li, Yan and Li, Xuanping and Chua, Tat-Seng},
  booktitle={Proceedings of the 29th ACM international conference on multimedia},
  pages={5382--5390},
  year={2021}
}

@inproceedings{park2009pairwise,
  title={Pairwise preference regression for cold-start recommendation},
  author={Park, Seung-Taek and Chu, Wei},
  booktitle={Proceedings of the third ACM conference on Recommender systems},
  pages={21--28},
  year={2009}
}

@inproceedings{lee2019melu,
  title={Melu: Meta-learned user preference estimator for cold-start recommendation},
  author={Lee, Hoyeop and Im, Jinbae and Jang, Seongwon and Cho, Hyunsouk and Chung, Sehee},
  booktitle={Proceedings of the 25th ACM SIGKDD international conference on knowledge discovery \& data mining},
  pages={1073--1082},
  year={2019}
}

@inproceedings{lu2020meta,
  title={Meta-learning on heterogeneous information networks for cold-start recommendation},
  author={Lu, Yuanfu and Fang, Yuan and Shi, Chuan},
  booktitle={Proceedings of the 26th ACM SIGKDD international conference on knowledge discovery \& data mining},
  pages={1563--1573},
  year={2020}
}

@article{vartak2017meta,
  title={A meta-learning perspective on cold-start recommendations for items},
  author={Vartak, Manasi and Thiagarajan, Arvind and Miranda, Conrado and Bratman, Jeshua and Larochelle, Hugo},
  journal={Advances in neural information processing systems},
  volume={30},
  year={2017}
}

@inproceedings{liu2020heterogeneous,
  title={A heterogeneous graph neural model for cold-start recommendation},
  author={Liu, Siwei and Ounis, Iadh and Macdonald, Craig and Meng, Zaiqiao},
  booktitle={Proceedings of the 43rd international ACM SIGIR conference on research and development in information retrieval},
  pages={2029--2032},
  year={2020}
}

@inproceedings{togashi2021alleviating,
  title={Alleviating cold-start problems in recommendation through pseudo-labelling over knowledge graph},
  author={Togashi, Riku and Otani, Mayu and Satoh, Shin'ichi},
  booktitle={Proceedings of the 14th ACM international conference on web search and data mining},
  pages={931--939},
  year={2021}
}

@inproceedings{huang2023aligning,
  title={Aligning distillation for cold-start item recommendation},
  author={Huang, Feiran and Wang, Zefan and Huang, Xiao and Qian, Yufeng and Li, Zhetao and Chen, Hao},
  booktitle={Proceedings of the 46th International ACM SIGIR Conference on Research and Development in Information Retrieval},
  pages={1147--1157},
  year={2023}
}

@inproceedings{zhou2023contrastive,
  title={Contrastive collaborative filtering for cold-start item recommendation},
  author={Zhou, Zhihui and Zhang, Lilin and Yang, Ning},
  booktitle={Proceedings of the ACM Web Conference 2023},
  pages={928--937},
  year={2023}
}

@article{cai2023user,
  title={User cold-start recommendation via inductive heterogeneous graph neural network},
  author={Cai, Desheng and Qian, Shengsheng and Fang, Quan and Hu, Jun and Xu, Changsheng},
  journal={ACM Transactions on Information Systems},
  volume={41},
  number={3},
  pages={1--27},
  year={2023},
  publisher={ACM New York, NY}
}

@inproceedings{dong2020mamo,
  title={Mamo: Memory-augmented meta-optimization for cold-start recommendation},
  author={Dong, Manqing and Yuan, Feng and Yao, Lina and Xu, Xiwei and Zhu, Liming},
  booktitle={Proceedings of the 26th ACM SIGKDD international conference on knowledge discovery \& data mining},
  pages={688--697},
  year={2020}
}

@inproceedings{yu2021personalized,
  title={Personalized adaptive meta learning for cold-start user preference prediction},
  author={Yu, Runsheng and Gong, Yu and He, Xu and Zhu, Yu and Liu, Qingwen and Ou, Wenwu and An, Bo},
  booktitle={Proceedings of the AAAI conference on artificial intelligence},
  volume={35},
  number={12},
  pages={10772--10780},
  year={2021}
}

@inproceedings{bi2020dcdir,
  title={DCDIR: A deep cross-domain recommendation system for cold start users in insurance domain},
  author={Bi, Ye and Song, Liqiang and Yao, Mengqiu and Wu, Zhenyu and Wang, Jianming and Xiao, Jing},
  booktitle={Proceedings of the 43rd international ACM SIGIR conference on research and development in information retrieval},
  pages={1661--1664},
  year={2020}
}

@inproceedings{finn2017model,
  title={Model-agnostic meta-learning for fast adaptation of deep networks},
  author={Finn, Chelsea and Abbeel, Pieter and Levine, Sergey},
  booktitle={International conference on machine learning},
  pages={1126--1135},
  year={2017},
  organization={PMLR}
}

@article{ha2016hypernetworks,
  title={Hypernetworks},
  author={Ha, David and Dai, Andrew and Le, Quoc V},
  journal={arXiv preprint arXiv:1609.09106},
  year={2016}
}

@article{chauhan2024brief,
  title={A brief review of hypernetworks in deep learning},
  author={Chauhan, Vinod Kumar and Zhou, Jiandong and Lu, Ping and Molaei, Soheila and Clifton, David A},
  journal={Artificial Intelligence Review},
  volume={57},
  number={9},
  pages={250},
  year={2024},
  publisher={Springer}
}

@inproceedings{beck2023hypernetworks,
  title={Hypernetworks in meta-reinforcement learning},
  author={Beck, Jacob and Jackson, Matthew Thomas and Vuorio, Risto and Whiteson, Shimon},
  booktitle={Conference on Robot Learning},
  pages={1478--1487},
  year={2023},
  organization={PMLR}
}

@inproceedings{radford2021learning,
  title={Learning transferable visual models from natural language supervision},
  author={Radford, Alec and Kim, Jong Wook and Hallacy, Chris and Ramesh, Aditya and Goh, Gabriel and Agarwal, Sandhini and Sastry, Girish and Askell, Amanda and Mishkin, Pamela and Clark, Jack and others},
  booktitle={International conference on machine learning},
  pages={8748--8763},
  year={2021},
  organization={PmLR}
}

@article{zhang2018graph,
  title={Graph hypernetworks for neural architecture search},
  author={Zhang, Chris and Ren, Mengye and Urtasun, Raquel},
  journal={arXiv preprint arXiv:1810.05749},
  year={2018}
}

@article{knyazev2021parameter,
  title={Parameter prediction for unseen deep architectures},
  author={Knyazev, Boris and Drozdzal, Michal and Taylor, Graham W and Romero Soriano, Adriana},
  journal={Advances in Neural Information Processing Systems},
  volume={34},
  pages={29433--29448},
  year={2021}
}

@inproceedings{zhmoginov2022hypertransformer,
  title={Hypertransformer: Model generation for supervised and semi-supervised few-shot learning},
  author={Zhmoginov, Andrey and Sandler, Mark and Vladymyrov, Maksym},
  booktitle={International Conference on Machine Learning},
  pages={27075--27098},
  year={2022},
  organization={PMLR}
}

@inproceedings{knyazev2023can,
  title={Can we scale transformers to predict parameters of diverse imagenet models?},
  author={Knyazev, Boris and Hwang, Doha and Lacoste-Julien, Simon},
  booktitle={International Conference on Machine Learning},
  pages={17243--17259},
  year={2023},
  organization={PMLR}
}

@inproceedings{zhang2024metadiff,
  title={Metadiff: Meta-learning with conditional diffusion for few-shot learning},
  author={Zhang, Baoquan and Luo, Chuyao and Yu, Demin and Li, Xutao and Lin, Huiwei and Ye, Yunming and Zhang, Bowen},
  booktitle={Proceedings of the AAAI conference on artificial intelligence},
  volume={38},
  number={15},
  pages={16687--16695},
  year={2024}
}

@article{schurholt2022hyper,
  title={Hyper-representations as generative models: Sampling unseen neural network weights},
  author={Sch{\"u}rholt, Konstantin and Knyazev, Boris and Gir{\'o}-i-Nieto, Xavier and Borth, Damian},
  journal={Advances in Neural Information Processing Systems},
  volume={35},
  pages={27906--27920},
  year={2022}
}

@article{shinn2023reflexion,
  title={Reflexion: Language agents with verbal reinforcement learning},
  author={Shinn, Noah and Cassano, Federico and Gopinath, Ashwin and Narasimhan, Karthik and Yao, Shunyu},
  journal={Advances in Neural Information Processing Systems},
  volume={36},
  pages={8634--8652},
  year={2023}
}

@article{wei2022chain,
  title={Chain-of-thought prompting elicits reasoning in large language models},
  author={Wei, Jason and Wang, Xuezhi and Schuurmans, Dale and Bosma, Maarten and Xia, Fei and Chi, Ed and Le, Quoc V and Zhou, Denny and others},
  journal={Advances in neural information processing systems},
  volume={35},
  pages={24824--24837},
  year={2022}
}

@inproceedings{hou2024llmrank,
  title={Large Language Models are Zero-Shot Rankers for Recommender Systems},
  author={Yupeng Hou and Junjie Zhang and Zihan Lin and Hongyu Lu and Ruobing Xie and Julian McAuley and Wayne Xin Zhao},
  booktitle={{ECIR}},
  year={2024}
}

@inproceedings{ni2022sentence,
  title={Sentence-t5: Scalable sentence encoders from pre-trained text-to-text models},
  author={Ni, Jianmo and Abrego, Gustavo Hernandez and Constant, Noah and Ma, Ji and Hall, Keith and Cer, Daniel and Yang, Yinfei},
  booktitle={Findings of the association for computational linguistics: ACL 2022},
  pages={1864--1874},
  year={2022}
}

@inproceedings{zhang2025llmtreerec,
  title={Llmtreerec: Unleashing the power of large language models for cold-start recommendations},
  author={Zhang, Wenlin and Wu, Chuhan and Li, Xiangyang and Wang, Yuhao and Dong, Kuicai and Wang, Yichao and Dai, Xinyi and Zhao, Xiangyu and Guo, Huifeng and Tang, Ruiming},
  booktitle={Proceedings of the 31st International Conference on Computational Linguistics},
  pages={886--896},
  year={2025}
}

@article{schurholt2021self,
  title={Self-supervised representation learning on neural network weights for model characteristic prediction},
  author={Sch{\"u}rholt, Konstantin and Kostadinov, Dimche and Borth, Damian},
  journal={Advances in Neural Information Processing Systems},
  volume={34},
  pages={16481--16493},
  year={2021}
}

@article{soro2024diffusion,
  title={Diffusion-based neural network weights generation},
  author={Soro, Bedionita and Andreis, Bruno and Lee, Hayeon and Jeong, Wonyong and Chong, Song and Hutter, Frank and Hwang, Sung Ju},
  journal={arXiv preprint arXiv:2402.18153},
  year={2024}
}

@article{xie2024weight,
  title={Weight Diffusion for Future: Learn to Generalize in Non-Stationary Environments},
  author={Xie, Mixue and Li, Shuang and Xie, Binhui and Liu, Chi and Liang, Jian and Sun, Zixun and Feng, Ke and Zhu, Chengwei},
  journal={Advances in Neural Information Processing Systems},
  volume={37},
  pages={6367--6392},
  year={2024}
}

@inproceedings{hou2022towards,
  title={Towards universal sequence representation learning for recommender systems},
  author={Hou, Yupeng and Mu, Shanlei and Zhao, Wayne Xin and Li, Yaliang and Ding, Bolin and Wen, Ji-Rong},
  booktitle={Proceedings of the 28th ACM SIGKDD Conference on Knowledge Discovery and Data Mining},
  pages={585--593},
  year={2022}
}

@inproceedings{li2023text,
  title={Text is all you need: Learning language representations for sequential recommendation},
  author={Li, Jiacheng and Wang, Ming and Li, Jin and Fu, Jinmiao and Shen, Xin and Shang, Jingbo and McAuley, Julian},
  booktitle={Proceedings of the 29th ACM SIGKDD Conference on Knowledge Discovery and Data Mining},
  pages={1258--1267},
  year={2023}
}

@inproceedings{qiu2021u,
  title={U-BERT: Pre-training user representations for improved recommendation},
  author={Qiu, Zhaopeng and Wu, Xian and Gao, Jingyue and Fan, Wei},
  booktitle={Proceedings of the AAAI Conference on Artificial Intelligence},
  volume={35},
  number={5},
  pages={4320--4327},
  year={2021}
}

@inproceedings{yao2022reprbert,
  title={ReprBERT: distilling BERT to an efficient representation-based relevance model for e-commerce},
  author={Yao, Shaowei and Tan, Jiwei and Chen, Xi and Zhang, Juhao and Zeng, Xiaoyi and Yang, Keping},
  booktitle={Proceedings of the 28th ACM SIGKDD Conference on Knowledge Discovery and Data Mining},
  pages={4363--4371},
  year={2022}
}

@inproceedings{zhang2022gbert,
  title={GBERT: Pre-training user representations for ephemeral group recommendation},
  author={Zhang, Song and Zheng, Nan and Wang, Danli},
  booktitle={Proceedings of the 31st ACM International Conference on Information \& Knowledge Management},
  pages={2631--2639},
  year={2022}
}

@article{gao2023chat,
  title={Chat-rec: Towards interactive and explainable llms-augmented recommender system},
  author={Gao, Yunfan and Sheng, Tao and Xiang, Youlin and Xiong, Yun and Wang, Haofen and Zhang, Jiawei},
  journal={arXiv preprint arXiv:2303.14524},
  year={2023}
}

@inproceedings{hou2024large,
  title={Large language models are zero-shot rankers for recommender systems},
  author={Hou, Yupeng and Zhang, Junjie and Lin, Zihan and Lu, Hongyu and Xie, Ruobing and McAuley, Julian and Zhao, Wayne Xin},
  booktitle={European Conference on Information Retrieval},
  pages={364--381},
  year={2024},
  organization={Springer}
}

@article{liu2023chatgpt,
  title={Is chatgpt a good recommender? a preliminary study},
  author={Liu, Junling and Liu, Chao and Zhou, Peilin and Lv, Renjie and Zhou, Kang and Zhang, Yan},
  journal={arXiv preprint arXiv:2304.10149},
  year={2023}
}

@inproceedings{sun2023chatgpt,
  title={Is ChatGPT Good at Search? Investigating Large Language Models as Re-Ranking Agents},
  author={Sun, Weiwei and Yan, Lingyong and Ma, Xinyu and Wang, Shuaiqiang and Ren, Pengjie and Chen, Zhumin and Yin, Dawei and Ren, Zhaochun},
  booktitle={EMNLP 2023}
}

@article{comanici2025gemini,
  title={Gemini 2.5: Pushing the frontier with advanced reasoning, multimodality, long context, and next generation agentic capabilities},
  author={Comanici, Gheorghe and Bieber, Eric and Schaekermann, Mike and Pasupat, Ice and Sachdeva, Noveen and Dhillon, Inderjit and Blistein, Marcel and Ram, Ori and Zhang, Dan and Rosen, Evan and others},
  journal={arXiv preprint arXiv:2507.06261},
  year={2025}
}

@article{hurst2024gpt,
  title={Gpt-4o system card},
  author={Hurst, Aaron and Lerer, Adam and Goucher, Adam P and Perelman, Adam and Ramesh, Aditya and Clark, Aidan and Ostrow, AJ and Welihinda, Akila and Hayes, Alan and Radford, Alec and others},
  journal={arXiv preprint arXiv:2410.21276},
  year={2024}
}

@inproceedings{dai2023uncovering,
  title={Uncovering chatgpt’s capabilities in recommender systems},
  author={Dai, Sunhao and Shao, Ninglu and Zhao, Haiyuan and Yu, Weijie and Si, Zihua and Xu, Chen and Sun, Zhongxiang and Zhang, Xiao and Xu, Jun},
  booktitle={Proceedings of the 17th ACM Conference on Recommender Systems},
  pages={1126--1132},
  year={2023}
}

@article{ma2024triple,
  title={Triple modality fusion: Aligning visual, textual, and graph data with large language models for multi-behavior recommendations},
  author={Ma, Luyi and Li, Xiaohan and Fan, Zezhong and Zhao, Kai and Xu, Jianpeng and Cho, Jason and Kanumala, Praveen and Nag, Kaushiki and Kumar, Sushant and Achan, Kannan},
  journal={arXiv preprint arXiv:2410.12228},
  year={2024}
}

@inproceedings{geng2022recommendation,
  title={Recommendation as language processing (rlp): A unified pretrain, personalized prompt \& predict paradigm (p5)},
  author={Geng, Shijie and Liu, Shuchang and Fu, Zuohui and Ge, Yingqiang and Zhang, Yongfeng},
  booktitle={Proceedings of the 16th ACM Conference on Recommender Systems},
  pages={299--315},
  year={2022}
}

@article{cui2022m6,
  title={M6-rec: Generative pretrained language models are open-ended recommender systems},
  author={Cui, Zeyu and Ma, Jianxin and Zhou, Chang and Zhou, Jingren and Yang, Hongxia},
  journal={arXiv preprint arXiv:2205.08084},
  year={2022}
}

@inproceedings{tan2024idgenrec,
  title={IDGenRec: LLM-RecSys Alignment with Textual ID Learning},
  author={Tan, Juntao and Xu, Shuyuan and Hua, Wenyue and Ge, Yingqiang and Li, Zelong and Zhang, Yongfeng},
  booktitle={Proceedings of the 47th International ACM SIGIR Conference on Research and Development in Information Retrieval},
  pages={355--364},
  year={2024}
}

@article{liao2023llara,
  title={Llara: Aligning large language models with sequential recommenders},
  author={Liao, Jiayi and Li, Sihang and Yang, Zhengyi and Wu, Jiancan and Yuan, Yancheng and Wang, Xiang and He, Xiangnan},
  journal={arXiv preprint arXiv:2312.02445},
  year={2023}
}

@inproceedings{he2017neural,
  title={Neural collaborative filtering},
  author={He, Xiangnan and Liao, Lizi and Zhang, Hanwang and Nie, Liqiang and Hu, Xia and Chua, Tat-Seng},
  booktitle={Proceedings of the 26th international conference on world wide web},
  pages={173--182},
  year={2017}
}

@inproceedings{acharya2023llm,
  title={Llm based generation of item-description for recommendation system},
  author={Acharya, Arkadeep and Singh, Brijraj and Onoe, Naoyuki},
  booktitle={Proceedings of the 17th ACM conference on recommender systems},
  pages={1204--1207},
  year={2023}
}

@article{agrawal2025gepa,
  title={Gepa: Reflective prompt evolution can outperform reinforcement learning},
  author={Agrawal, Lakshya A and Tan, Shangyin and Soylu, Dilara and Ziems, Noah and Khare, Rishi and Opsahl-Ong, Krista and Singhvi, Arnav and Shandilya, Herumb and Ryan, Michael J and Jiang, Meng and others},
  journal={arXiv preprint arXiv:2507.19457},
  year={2025}
}

@inproceedings{WangYXSDZ22,
  author    = {Dong Wang and Shaoguang Yan and Yunqing Xia and Kav{\'e} Salamatian and Weiwei Deng and Qi Zhang},
  title     = {Learning Supplementary NLP Features for CTR Prediction in Sponsored Search},
  booktitle = {Proceedings of the 28th ACM SIGKDD Conference on Knowledge Discovery and Data Mining (KDD '22)},
  pages     = {4010--4020},
  year      = {2022},
  publisher = {ACM},
  address   = {Washington, DC, USA},
  doi       = {10.1145/3534678.3539064}
}

@inproceedings{jiang2025beyond,
  title={Beyond Utility: Evaluating LLM as Recommender},
  author={Jiang, Chumeng and Wang, Jiayin and Ma, Weizhi and Clarke, Charles LA and Wang, Shuai and Wu, Chuhan and Zhang, Min},
  booktitle={Proceedings of the ACM on Web Conference 2025},
  pages={3850--3862},
  year={2025}
}

@article{song2024importance,
  title={The importance of online data: Understanding preference fine-tuning via coverage},
  author={Song, Yuda and Swamy, Gokul and Singh, Aarti and Bagnell, J and Sun, Wen},
  journal={Advances in Neural Information Processing Systems},
  volume={37},
  pages={12243--12270},
  year={2024}
}

@inproceedings{yan2022scale,
  title={Scale calibration of deep ranking models},
  author={Yan, Le and Qin, Zhen and Wang, Xuanhui and Bendersky, Michael and Najork, Marc},
  booktitle={Proceedings of the 28th ACM SIGKDD Conference on Knowledge Discovery and Data Mining},
  pages={4300--4309},
  year={2022}
}

@inproceedings{mcmahan2013ad,
  title={Ad click prediction: a view from the trenches},
  author={McMahan, H Brendan and Holt, Gary and Sculley, David and Young, Michael and Ebner, Dietmar and Grady, Julian and Nie, Lan and Phillips, Todd and Davydov, Eugene and Golovin, Daniel and others},
  booktitle={Proceedings of the 19th ACM SIGKDD international conference on Knowledge discovery and data mining},
  pages={1222--1230},
  year={2013}
}

@inproceedings{Liang2020JTCN,
  author    = {Tingting Liang and Congying Xia and Yuyu Yin and Philip S. Yu},
  title     = {Joint Training Capsule Network for Cold Start Recommendation},
  booktitle = {Proceedings of the 43rd International ACM SIGIR Conference on Research and Development in Information Retrieval (SIGIR ’20)},
  year      = {2020},
  organization = {ACM},
  note      = {arXiv:2005.11467},
  url       = {https://arxiv.org/abs/2005.11467}
}

@inproceedings{cheng2016wide,
  title={Wide \& deep learning for recommender systems},
  author={Cheng, Heng-Tze and Koc, Levent and Harmsen, Jeremiah and Shaked, Tal and Chandra, Tushar and Aradhye, Hrishi and Anderson, Glen and Corrado, Greg and Chai, Wei and Ispir, Mustafa and others},
  booktitle={Proceedings of the 1st workshop on deep learning for recommender systems},
  pages={7--10},
  year={2016}
}

@article{Roustaei2024LinearRegression,
  author    = {Narges Roustaei},
  title     = {Application and Interpretation of Linear-Regression Analysis},
  journal   = {Medical Hypothesis, Discovery \& Innovation in Ophthalmology},
  volume    = {13},
  number    = {3},
  pages     = {151--159},
  year      = {2024},
  doi       = {10.51329/mehdiophthal1506},
  url       = {https://pubmed.ncbi.nlm.nih.gov/39507810}
}

\appendix
\clearpage

\section{LLM-HYPER Implementation Details and Trade-offs}
\label{app:llm_hyper_implement}

All LLM-HYPER variants perform weight generation offline before computing the CTR for ranking and latency evaluation. 

When generating the weights, we explored various configurations of user feature combinations to improve generation. While asking LLMs to generate one feature weight at a time maintains high tractability, it loses cross-feature connections for weight reasoning and is very costly (in time and money). However, providing the definitions of all features simultaneously reduces instructional following and increases the risk of hallucination. Thus, the feature batch size is very critical for successful weight generations. 
We find that 5 to 10 features at a time provide good cross-feature connectivity and tractability for LLM generation. 

Generation time varies across LLM backbones. Table \ref{tab:generation-time} summarizes the average generation time in seconds for an ad. Combined with Table \ref{tab:model-performance}, we can see that Gemini-2.5-Pro offers the best performance in CTR ranking but also takes the longest generation time offline. Although the long generation time won't affect the online serving, Gemini-2.5-Pro is still more expensive than its lighter version, Gemini-2.5-Flash. This provides a trade-off for backend LLMs to balance overall LLM usage and serving performance. 

\begin{table}[h!]
\centering
\begin{tabular}{lc}
\toprule
\textbf{Models} & \textbf{Generation Time(s)} \\ 
\midrule
\textbf{GPT-4o}  & 47.7 \\ 
\textbf{GPT-5.1}  & 58.7 \\ 
\textbf{Gemini-2.5-Flash} &   83.3 \\
\textbf{Gemini-2.5-Pro} &   102.9 \\
\bottomrule
\end{tabular}
\caption{Offline LLM weight generation time (averaged by each ad)}
\label{tab:generation-time}
\end{table}

\section{Metrics Definition for Explainability} 
\label{app:appendix_metrics}

To rigorously evaluate the explainability and alignment of \textbf{LLM-HYPER}, we define three primary metrics: \textit{Hitrate@5}, \textit{Coverage@5}, and \textit{Consistency Rate}. These metrics quantify LLM-Hyper's ability to map ad to feature-level preferences and ensure LLM's internal reasoning and prediction coherence.

\paragraph{Hitrate@5 (HR@5)}
This metric measures whether the ground-truth target feature is identified within LLM's top-5 predictions. For each ad $i$, let $P_{gt,i}$ be the target feature and $\mathcal{P}_{top5,i}$ be the set of top-5 features ranked by their predicted weights in descending order. The Hitrate@5 is defined as:
\begin{equation}
\text{HR@5} = \frac{1}{M} \sum_{i=1}^{M} \mathbb{I}(P_{gt,i} \in \mathcal{P}_{top5,i})
\end{equation}
where $M$ denotes the total number of evaluation ad and $\mathbb{I}(\cdot)$ is the indicator function.

\paragraph{Coverage@5}
To assess the semantic overlap between the predicted weights and the ground-truth features, we employ a Jaccard-based coverage metric. For cases where an ad align with multiple features (represented as a set $\mathcal{P}_{gt,i}$), Coverage@5 quantifies the intersection over union (IoU) between the predicted top-5 set $\mathcal{P}_{top5,i}$ and the ground-truth set:
\begin{equation}
\text{Coverage@5} = \frac{1}{M} \sum_{i=1}^{M} \frac{|\mathcal{P}_{top5,i} \cap \mathcal{P}_{gt,i}|}{|\mathcal{P}_{top5,i} \cup \mathcal{P}_{gt,i}|}
\end{equation}
Specifically, in scenarios where $\mathcal{P}_{gt,i}$ contains only a single target feature, this metric reflects the precision of the LLM's high-confidence alignment.

\paragraph{Consistency Rate}
Consistency Rate evaluates the logical alignment between the LLM's natural language reasoning ($R_i$) and its generated numerical weight scores ($S_i$). We utilize \textbf{Sign Agreement} as the primary measure. An auxiliary \textit{LLM-Judge} maps the reasoning text $R_i$ into a sentiment polarity $\text{sign}(R_i) \in \{\text{positive}, \text{neutral}, \text{negative}\}$. Similarly, the predicted score $S_i$ is mapped based on its scalar value (relative to a threshold of $0$). The Consistency Rate is calculated as:
\begin{equation}
\text{Consistency Rate} = \frac{1}{M} \sum_{i=1}^{M} \mathbb{I}(\text{sign}(R_i) = \text{sign}(S_i))
\end{equation}
In practice, we utilize Gemini-2.5-Pro as the \textit{LLM-Judge} to provide the sentiment polarity sign.

\section{Deployment Insights}
\label{app:deployment_insights}
In this section, we discuss insights into deployment.

\subsection{Reduction of Hallucination and Randomness}
Hallucinations and randomness in LLM outputs could slow down weight initialization due to incorrect information and improper numeric scale, respectively. 
For example, hallucinated output could create a non-existing connection, misleading the weight generation and simulating incorrect user-ad interactions. Randomness affects the robustness, especially for features with borderline importance. 

During deployment, we take several practices to reduce hallucinations and randomness. 
First, we ask LLM to generate both reasoning and weight to enforce the self-reflection. Figure \ref{fig:full_prompt} and \ref{fig:prompt_score_reasoning} show the emphasis of reasoning.
Second, we employ the LLM-as-judge technique to validate the generation.
The incorrect information will be updated after regeneration. 
These two help reduce hallucination effectively. 
To reduce randomness, especially in contextualized and borderline features, we ask the LLM to generate the output multiple times at the given budget and take the average, thereby reducing randomness.
Furthermore, the mini-batch-style feature-weight prediction in Appendix \ref{app:llm_hyper_implement} provides additional context, thereby stabilizing weight generation for borderline features. 

\subsection{Regulation of Improper Contents}
Improper content could happen if the cold-start ranker ranks the improper ad to users.
For example, ranking a meat-related ad to a user with a strong preference for a vegetarian diet not only harms ranking metrics but also diminishes the user experience. 
Because the reasoning and weights are generated by LLM-HYPER, we use the same LLM-as-judge technique to cross-validate the feature definition and ad content. This check serves as a filter rather than simply assessing the correctness of scores, meaning that filtered ads will be tagged and removed from the ad candidates for ranking and serving if a user has a high value for the corresponding features. 
Because we only need to validate the generation per ad rather than per user, the check can be performed efficiently at a manageable LLM cost. 

\subsection{Prompt Optimization}
The prompt for weight generation in LLM-HYPER is further optimized in practice using DSPy\footnote{https://dspy.ai/} and GEPA~\cite{agrawal2025gepa}, incorporating human rater feedback and instructions from the trained model for warm-start ads.

\section{Visualization of LLM-generated Weights}
\label{app:llm_hyper_result_visualiation}
In ad ranking, many input features are clearly defined. For example, engagement-related features are usually defined by the number of clicks and view events. 
We use an example of the engagement metric defined for toy- and game-related ads— \texttt{feature\_toys\_games}: \textit{Customer engagement with toys, games, and puzzle categories}.
LLM-HYPER will generate weights based on the ad's content and its relevance to this feature definition. 
Table \ref{tab:feature_weights} summarizes the visualized results, including the LLM reasoning along with the generated weights. 
LLM-HYPER generates the weight value based on the ad content. 
For Ad 1, which is about toy and construction sets, LLM-HYPER assigns a large weight value  due to its strong connection. 
For Ad 2, which is about outdoor furniture, LLM-HYPE generates a very low weight value because of the conflicting relationship between toys and furniture. 
For Ad 3, LLM-HYPER predicts a neutral score due to weak relevance. 
These results indicate that LLM-HYPER can translate LLM's strong reasoning capability into linear model weights to simulate user-ad interaction, mitigating the cold-start issue.

\begin{table*}[t]
\centering
\begin{tabular}{p{0.8cm}|p{3cm}|p{1.2cm}|p{3.5cm}|p{5cm}}
\toprule
\textbf{Ad} & \textbf{Ad Content} & \textbf{Weight} & \textbf{Key Signals} & \textbf{LLM Rationale} \\ \hline

Ad 1 & 
Age-appropriate building toys and construction sets & 
0.95 & 
\textbullet~Strong: Toy and game category engagement\newline
\textbullet~Strong: Building and construction toy affinity & 
The dominant feature activation centers on toy and building set interactions. The Ad's product category and thematic content closely align with this feature's predictive pattern, indicating strong relevance for engagement prediction. \\ \hline

Ad 2 & 
High-value outdoor furniture with minor recreational overlap & 
-0.35 & 
\textbullet~Strong: Indoor recreational products\newline
\textbullet~Weak: Outdoor lawn activities & 
The dominant feature signal centers on indoor play items. The secondary outdoor activity signal is insufficient to predict interest in premium outdoor living categories. \\ \hline

Ad 3 & 
Children's footwear (Crocs) with shared toy purchaser demographic & 
0.0 & 
\textbullet~Weak: Children's product category overlap\newline
\textbullet~Weak: Parent shopper demographic & 
The feature captures toy purchasing behavior, which correlates with parents who also purchase children's apparel and footwear. The cross-category purchasing pattern creates a neutral signal for children's fashion items despite product category mismatch. \\ 
\bottomrule

\end{tabular}
\caption{Feature weight predictions for \texttt{feature\_toys\_games} feature across different ad contexts.}
\label{tab:feature_weights}
\end{table*}

\section{More Evaluation Details: Counterfactual Robustness Evaluation}
\label{app:counterfactual}

To evaluate the robustness of \textbf{LLM-HYPER} in industrial settings where ground-truth interaction labels are frequently unavailable, we introduce a counterfactual experiment. This experiment assesses whether the model consistently adjusts its weight generation in response to controlled semantic perturbations.

\subsection{Perturbation Taxonomy and Construction}
For each ad, we apply three distinct types of semantic modifications to the ad title and summary using an auxiliary LLM generator. To ensure semantic validity, these modifications are conditioned on the human-labeled \textit{target feature}:

\begin{itemize}
    \item \textbf{Enhanced}: Rewrites the ad to strengthen its alignment with the target feature's latent interests. 
    \item \textbf{Diminished}: Rewrites the ad to conflict with or deviate from the target feature.
    \item \textbf{Neutralized}: Rewrites the ad to remove specific feature-targeting cues, resulting in a generic description.
\end{itemize}

The prompt used for Counterfactual modification is shown in Figure \ref{fig:prompt_counterfactual}. 

\subsection{Metric: Accuracy of Weights Direction}

To evaluate the \textbf{LLM-HYPER}'s robustness, we formulate a binary classification task to determine whether \textbf{LLM-HYPER} adjusts the generated weights in the same semantic direction as the applied modifications. We define the \textbf{Accuracy of Weights Direction} as the primary metric. 

For each modification type $m \in \{\text{Enhanced, Diminished, Neutralized}\}$, let $S_{\text{orig}}$ denote the weight of the original ad and $S_{\text{cf}}$ denote the weight of the counterfactual ad. For the $i$-th sample, the binary prediction is considered correct ($D(i, m) = 1$) if the shift in weight aligns with the semantic intent of the modification:

\begin{equation}
D(i, m) =
\begin{cases}
1 & \text{if } \begin{aligned} 
&(m = \text{Enhanced} \wedge S_{\text{cf}} > S_{\text{orig}}) \ \vee \\
&(m = \text{Diminished} \wedge S_{\text{cf}} < S_{\text{orig}}) \ \vee \\
&(m = \text{Neutralized} \wedge |S_{\text{cf}}| < |S_{\text{orig}}|) 
\end{aligned} \\
0 & \text{otherwise}
\end{cases}
\end{equation}

The aggregate Accuracy for this binary classification task is calculated as:
\begin{equation}
\text{Accuracy} = \frac{1}{N} \sum_{i=1}^{N} D(i, m)
\end{equation}
where $N$ represents the total number of counterfactual samples. This metric quantifies the LLM's ability to reliably translate semantic perturbations into consistent numerical weight adjustments for the human-labeled features.
\subsection{Semantic Perturbation Examples}
Table~\ref{tab:cf_comprehensive} provides examples of the ad transformations across different modification settings used in our robustness experiments.

\begin{table*}[t]
\centering
\begin{tabular}{p{2cm}|p{2cm}|p{2cm}|p{3.5cm}|p{1cm}|p{1cm}|p{1.5cm}}
\toprule
\textbf{Original Ad} & \textbf{Target Features} & \textbf{Mod Type} & \textbf{Modified Ad} & \textbf{$S_{\text{orig}}$} & \textbf{$S_{\text{cf}}$} & \textbf{$\Delta$ Score} \\ \midrule
Patio \& Garden & outdoor activity \& sports & \textit{Enhanced} & \textbf{Game Day Upgrades: Patio \& Grill} & 0.192 & 0.735 & $+0.543$ \\ \hline
Vitamins \& Supps & Healthy Living & \textit{Neutralized} & \textbf{Daily Essentials} & 1.055 & 0.975 & $-0.080$ \\ \hline
Fall Home Trends & Home Decor & \textit{Diminished} & \textbf{Fall Gutter \& Roof Maintenance} & 3.750 & 1.150 & $-2.600$ \\  
\bottomrule
\end{tabular}
\caption{Examples of counterfactual semantic modifications and the resulting weight changes generated by \textbf{LLM-HYPER} (\textbf{LH$_{2.5P}$}).}
\label{tab:cf_comprehensive}
\end{table*}

\subsection{Case Study: LLM Prediction on Counterfactual Samples }
To verify that \textbf{LLM-HYPER} is not merely reacting to keyword statistics but is performing grounded reasoning, we provide the following representative cases:

\paragraph{Case 1: Enhanced Sample.} For the ad \textit{``Fall Savings Patio And Garden''} targeted at a outdoor activity\&sports feature, the model initially identifies a positive relationship. Upon modification to \textit{``Game Day Upgrades: Patio \& Grill Savings''}, the model's internal reasoning successfully identifies \textit{``Grill''} and \textit{``Game Day''} as high-affinity features for the feature's interest in social sports gatherings. Consequently, the assigned weight increases from $0.192$ to $0.735$ ($\Delta = +0.543$).

\paragraph{Case 2: Neutralized Sample.} In the \textit{``Vitamins And Supplements''} case for the \textit{``Healthy Living''} feature, the original score ($1.055$) indicates a strong match. When neutralized to \textit{``Daily Essentials''}, the model's reasoning explicitly notes the removal of health-specific utility and the loss of medical-grade branding, leading to a dampened weight of $0.975$ ($\Delta = -0.080$), correctly reverting toward a neutral baseline.

\paragraph{Case 3: Diminished Sample.} For the ad \textit{``Fall Home Trends''} targeted at a \textit{``Home Decor''} feature, the original title aligns with interior aesthetics ($3.750$). However, modifying it to \textit{``Fall Gutter \& Roof Maintenance''} shifts the semantic focus to exterior structural labor. The LLM's reasoning highlights that this new focus lacks the decorative appeal expected by the feature, resulting in a sharp penalty to $1.150$ ($\Delta = -2.600$).

\section{Related Work}
\noindent \textbf{Cold-start Recommendation} has been studied in many past works \cite{wei2021contrastive, park2009pairwise, lee2019melu, lu2020meta, vartak2017meta}. Methods based on content features of cold items (customers) leverage the content similarity between warm and cold data for semantic relevance enhancement \cite{liu2020heterogeneous, togashi2021alleviating, cai2023user} and 
knowledge alignment \cite{wei2021contrastive, huang2023aligning, zhou2023contrastive} to enhance the learning on the incomplete labels. Other methods adapt the model agnostic meta-learning framework \cite{finn2017model} to  address the efficient learning on limited labels \cite{lee2019melu, vartak2017meta, dong2020mamo, yu2021personalized, bi2020dcdir}. However, all these solutions rely on training labels. Our solution overcomes this limitation in the strict cold start, where no labels are available during the training stage, and leverages LLMs to generate the model weight directly without training. 

\noindent \textbf{Hypernetwork} is a neural network that generates model weights for another neural network (i.e., target network) \cite{ha2016hypernetworks, chauhan2024brief}. Hypernetworks have shown promising results in a variety of deep learning problems \cite{beck2023hypernetworks, zhang2018graph, knyazev2021parameter, zhmoginov2022hypertransformer, knyazev2023can, zhang2024metadiff} and parameter generation from pretrained distributions has been addressed in \cite{schurholt2021self, schurholt2022hyper}. Recent works \cite{soro2024diffusion, xie2024weight} further push the boundary to explore diffusion-based solutions for unseen distributions. However, these solutions are not applicable in the strict cold start scenario due to the absence of training labels. Our proposed LLM-HYPER fills the gap by generating weights from the multi-modal description of the unseen distribution.

\noindent \textbf{Large Language Models}: applications in recommender systems fall into three categories: representation learning, in-context learning, and generative recommendation. In representation learning, LLMs enhance user and item representations by utilizing contextual data during the recommendation process~\cite{hou2022towards, li2023text, qiu2021u, yao2022reprbert, zhang2022gbert}. 
In-context learning, LLMs follow instructions to directly generate recommendations without the need for training or fine-tuning~\cite{gao2023chat, hou2024large, liu2023chatgpt, sun2023chatgpt, dai2023uncovering} even under cold start scenarios. 
LLM-based generative recommendation directly generates the target item in natural language \cite{geng2022recommendation, cui2022m6, liao2023llara, tan2024idgenrec, ma2024triple}.
However, running LLMs in the production environment could be costly. LLM-HYPER generates the model weights for effective real-time performance.

\section{LLM-HYPER Prompts Design}
\label{app:llm_hyper_prompt_design}
We provide the complete prompt template for the 3-shot CoT with the feature batch size of 5 in Figure \ref{fig:full_prompt}.
Note that the template and instructions could be modified according to actual ad ranking scenarios. We focus on e-commerce ad ranking, so we highlight e-commerce-related guidelines in the prompt template. 

LLM reasoning can be emphaizd by updating the output format instruction. Figure \ref{fig:prompt_score_reasoning} presents an enhanced version by adding multiple reasoning dimensions, such as context alignment detection, summarization of matching aspects and rationales. 

Many LLMs support multimodal reasoning. Besides image caption, the actual image file could be encoded for better reasoning. This practice will help the robust deployment in an industrial environment:
\begin{verbatim}
# Create message with an image
message = HumanMessage(
    content=[
        {"type": "text", "text": user_prompt},
        {
            "type": "image_url",
            "image_url": f"{base64_image}"
        }
    ]
)

# Get response
response = llm.invoke([message])
\end{verbatim}

\begin{figure*}[t]
\begin{lstlisting}[style=promptstyle]
<SYS>## System Prompt
You are a personalization scoring system that generates feature weights for predicting customer engagement with advertisements. Your role is to analyze customer preferences and ad content to determine the importance of different features in calculating relevance scores.</SYS>

<USER>## User Prompt
**Reference Examples:**
Below are 3 past ads with their corresponding weights to guide your predictions:

Example 1:
- Ad Title: {few_shot[0]['ad_title']}
- Image Summary: {few_shot[0]['image_caption']}
- Weights: {few_shot[0]['ad_weights']}

Example 2:
- Ad Title: {few_shot[1]['ad_title']}
- Image Summary: {few_shot[1]['image_caption']}
- Weights: {few_shot[1]['ad_weights']}

Example 3:
- Ad Title: {few_shot[2]['ad_title']}
- Image Summary: {few_shot[2]['image_caption']}
- Weights: {few_shot[2]['ad_weights']}



**Your task**
Given a customer with specific preferences and shopping behavior, generate feature weights for predicting the engagement score when this customer sees the following ad:

**Target Ad:**
Generate feature weights for the following ad:
- Ad Title: {input_data['ad_title']}
- Image Summary: {input_data['image_caption']}

**User Features:**
- Features: {input_data['features']}
- Descriptions: {input_data['feature_descriptions']}

**Instructions:**
1. Identify key customer interests from the profile
2. Analyze the target ad to determine product categories
3. Compare ad content with customer interests
4. Assess relevance and alignment
5. Generate weights reflecting each feature's contribution

**Guidelines:**
- Weights follow a normal distribution centered near zero
- Typical range: -1 to 1
- Consider similarity to reference examples
- Only exceed range if examples justify it</USER>

<FMT>## Output Format
Return ONLY a JSON object with 5 feature weights. No explanations.

{
  "feature_1": 0.123,
  "feature_2": -0.456,
  "feature_3": 0.789,
  "feature_4": -0.234,
  "feature_5": 0.567
}</FMT>
\end{lstlisting}
\caption{Prompt template for generating feature weights in LLM-HYPER.}
\label{fig:full_prompt}
\end{figure*}

\begin{figure*}[t]
\small
\begin{lstlisting}[style=promptstyle]
<SYS>## System Prompt
[same as the regular LLM-HYPER system prompt]</SYS>

<USER>## User Prompt
[same as the regular LLM-HYPER user prompts]</USER>

<FMT>## Formatting Requirements
Provide ONLY the JSON output:
{
  "ad_analysis": "what the ad offers",
  "alignment": "matches and mismatches",
  "key_factors": ["factor1", "factor2"],
  "reasoning_summary": "brief explanation",
  "predicted_score": 0.00
}</FMT>
\end{lstlisting}
\caption{Prompt for weight prediction with emphasis of chain-of-thought reasoning.}
\label{fig:prompt_score_reasoning}
\end{figure*}

\begin{figure*}[t]
\small
\begin{lstlisting}[style=promptstyle]
<SYS>## System Prompt
[same as the regular LLM-HYPER system prompt]</SYS>

<USER>## User Prompt
**Original Ad:**
- Title: {original_title}
- Image Summary: {original_summary}

**Target Features:**
- Features: {input_data['features']}
- Descriptions: {input_data['feature_descriptions']}

**Modification Type: {modification_type}**

For 'enhanced': Modify to BETTER align with target features
For 'diminished': Modify to LESS align with target features
For 'neutralized': Make more GENERIC, removing category references

Keep modifications realistic and subtle.</USER>

<FMT>## Formatting Requirements
Provide ONLY the JSON output:
{
  "modified_title": "new title",
  "modified_summary": "new summary",
  "modification_explanation": "rationale for changes"
}</FMT>
\end{lstlisting}
\caption{Prompt for generating counterfactual Ad modifications.}
\label{fig:prompt_counterfactual}
\end{figure*}

\end{document}